# DMLO: Deep Matching LiDAR Odometry

Zhichao Li[1] and Naiyan Wang[2]

*Abstract*— LiDAR odometry is a fundamental task for various areas such as robotics, autonomous driving. This problem is difficult since it requires the systems to be highly robust running in noisy real-world data. Existing methods are mostly local iterative methods. Feature-based global registration methods are not preferred since extracting accurate matching pairs in the nonuniform and sparse LiDAR data remains challenging. In this paper, we present Deep Matching LiDAR Odometry (DMLO), a novel learning-based framework which makes the feature matching method applicable to LiDAR odometry task. Unlike many recent learning-based methods, DMLO explicitly enforces geometry constraints in the framework. Specifically, DMLO decomposes the 6-DoF pose estimation into two parts, a learning-based matching network which provides accurate correspondences between two scans and rigid transformation estimation with a close-formed solution by Singular Value Decomposition (SVD). Comprehensive experimental results on real-world datasets KITTI and Argoverse demonstrate that our DMLO dramatically outperforms existing learning-based methods and comparable with the state-of-the-art geometry-based approaches.

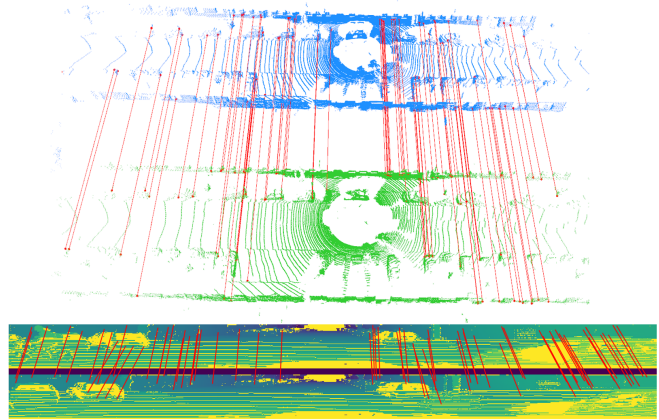

Fig. 1. The key idea of our proposed DMLO method is to convert the matching in sparse 3D space into (almost) dense 2D cylinder image. We show the top 100 inliners predicted by DMLO in cylinder image(bottom) and their corresponding 3D position in point cloud(top). Filtering by our confidence, there are almost no outliers in the matching.

## I. INTRODUCTION

Estimating ego-motion (a.k.a odometry) in dynamic and complicated real-world environments is a vital task for autonomous vehicles and robots. This task along with perception system are prerequisites for subsequent path planning tasks. These robots are often equipped with multiple sensors such as cameras and LiDARs to perceive the surrounding environment. Different from the camera, LiDAR not only can provide high-precision 3D measurements but also is insensitive to lighting conditions which makes it complementary for camera-based systems. However, LiDAR point clouds are sparse (usually consists of 32 or 64 lines) such that accurate ego-motion estimation is difficult in its nature.

Estimating frame to frame translation and rotation is fundamental in a LiDAR odometry system [1], [2]. Traditional methods can be roughly categorized into two types, iterative closest-based local registration [3], [4], [5], [2] and feature-based global registration [6], [7]. Feature-based methods first find a set of correspondences between two scans and then directly solve the best relative translation and rotation between them. However, due to the nonuniformity and sparsity of the LiDAR point clouds, these methods often fail to find enough high-precision matching pairs. Local methods are more popular in odometry task. ICP variants [3], [4] solve this problem by minimizing distances between closest points in consecutive scans. NDT [5] transforms the scan-to-scan problem to a point-to-distribution likelihood maximization task which makes the registration more robust. To prevent mismatches, LOAM [2] only aligns points in edge-line or plane-surface. Nevertheless, the problem becomes unfriendly for local methods when the pose variation is large and no initial alignment is given. Furthermore, it is computationally expensive, because it performs the nearest neighbor queries iteratively.

Recently, some deep learning based methods [8], [9], [10], [11], [12], [13] have excelled traditional methods in some 3D estimation tasks. Notably, LO-Net [14] is the first successful attempt toward learning-based LiDAR odometry. Nevertheless, LO-Net highly relies on the fitting power of CNN. It simply inputs two successive scans, and directly output the relative motion between them. The network is trained in an end-to-end manner, and no geometric constraints are imposed. Therefore, it is prone to over-fitting to the scene and not interpretable to the failure cases.

Traditional feature-based LiDAR odometry method suffers from inferior matching accuracy, while LO-Net cannot make full use of the geometric constraints. Is there a method could get the best of the world? Our answer is yes. The key idea of our proposed method is to utilize the property that one LiDAR scan is from a single viewpoint. Specifically, we could encode all the information to a projected 2D image without any information loss (as shown in Fig. 1). Based on this encoding, our method first extracts grid-wise feature vectors using CNN and compare the similarities in a local region to get correspondences between different scans. We additionally predict the confidence for every correspondence to ease the selection of accurate pairs, since only three pairs of matching is needed to solve the relative motion in 3D space. After that, the problem is converted to minimize the distances between all matched pairs in 3D space which can be simply solved by Singular Value Decomposition (SVD). We conduct detailed ablation studies to justify the efficacy of each component in our proposed approach. Finally, we evaluate our method on the standard KITTI benchmark [15]. *For the first time, we demonstrate global sparse matching is feasible for LiDAR odometry task.* We also test our proposed method on the recent Argoverse [16] dataset. The encouraging results further demonstrate the universality of our proposed method under different hardware configurations.

[1]Zhichao Li and [2]Naiyan Wang is with Tusimple, Beijing, China. {`leeisabug, winsty`}@gmail.com

To summarize, our contributions are three folds:

- We propose a deep learning approach to extract high confidence matching pairs from successive LiDAR scans.
- Based on this powerful matching and confidence estimation method, we propose a simple yet effective LiDAR odometry system named Deep Matching LiDAR Odometry (DMLO).
- We test our proposed method on standard KITTI and a recently proposed Argoverse dataset. Extensive results demonstrate that our method is on par or even better than traditional LiDAR odometry methods.

## II. RELATED WORK

In this section, we will briefly review some closely related work in LiDAR odometry area, then followed by some recent works that utilize deep learning techniques for 3D estimation tasks.

### A. Local Iterative Methods

Local iterative methods have a long-standing history in Simultaneous Localization and Mapping task (SLAM). The basic idea is to adjust the transformation iteratively based on local criteria fitting. The most classical method is Iterative Closest Point (ICP) [3], [17]. It begins with a rough initial alignment and iteratively estimates the optimal transformation by minimizing the distance between the closest point pairs. Despite its simplicity, it suffers from expensive computational cost and sensitivity to noise data. Various subsequent approaches [18], [19], [20], [21], [22], [23] are devised to solve these problems. Fitzgibbonet al.[24] introduced a robust kernel in ICP's optimization to increase the robustness. Bouazizet al.[25] used sparsity inducing norms to deal with outliers and incomplete data. Generalized-ICP (GICP) [4] unified point-to-point ICP and point-to-plane ICP into a single probabilistic framework and shown better performance. Instead of using explicit point to point correspondences, others also tried to design more robust hand-crafted criteria. Normal Distributions Transform(NDT) [5] first fits a multivariate normal distribution to every voxel in one scan. Then the registration problem is turned into finding a transformation to maximize the probability of the other scan. LiDAR Odometry And Mapping (LOAM) [2], [26] extracts line and plane features instead of point-to-point matching to reduce mismatches, and is currently the state-of-the-art LiDAR SLAM method.

Albeit these local iterative methods are widely used, a proper initialization close enough to global optima is essential. Nevertheless, in the case of abrupt motion, this assumption is usually violated, which may lead to inferior results. Some other methods [27], [28], [29], [30] introduce branch-and-bound (BnB) framework to guarantee global optimality. However, the computation complexity of full search is usually unaffordable.

### B. Global Sparse Registration

Global Sparse Representation is mainly based on local geometric descriptors[31], [32], Most of these works are based on hand-crafted features. Some representative works include Persistent Feature Histograms (PFH) [7], which describes the curvature around a point by estimating surface normals between neighbor points. Fast Point Feature Histograms (FPFH) [6] is the faster version of it and can be used in real-time applications. For these hand-crafted features, variants of RANSAC is typically used to reject outliers and then estimates the transformation matrix. Based on FPFH, Fast Global Registration(FGR) [33] solves an optimization problem with robust kernel which aims to suppress the outliers. However, the sparsity and nonuniformity of LiDAR point cloud make these hand-crafted representations prone to noise and variation of scenes.

In contrast to these existing LiDAR keypoint methods, our method directly calculates the matching between two scans in a local area. We don't need separate keypoint detector and descriptor into two stages. Moreover, our method could naturally output accurate confidence estimation, which enables the subsequent outlier rejection task.

### C. Deep-learning Methods

With the recent success of deep neural networks, CNN-based methods also demonstrate their potential in 3D computer vision applications. For example, [9], [8], [10] propose a new unsupervised visual odometry(VO) framework which trains the depth and ego-motion estimation simultaneously. For the point cloud registration, feature-based methods [34], [35], [36], [37] try to detect more accurate correspondences. The other methods [38], [39] follow the direction of iterative local methods which align two scans directly and outperform ICP variants. Compared with registration, odometry problems are more challenging which requires much higher accuracy under real noisy environments to prevent drifting. In particular, Velaset al. [40] predicted pose using classification by exhausting all possibilities, while LO-net [14] directly formulates the odometry estimation as a numerical regression. Both these works are not geometric interpretable, thus the generalization on different configurations and scenes cannot be guaranteed.

## III. METHODS

In this section, we first give a brief overview of our proposed Deep Matching Odometry (DMLO) method, and then elaborate on each component in subsequent subsections.

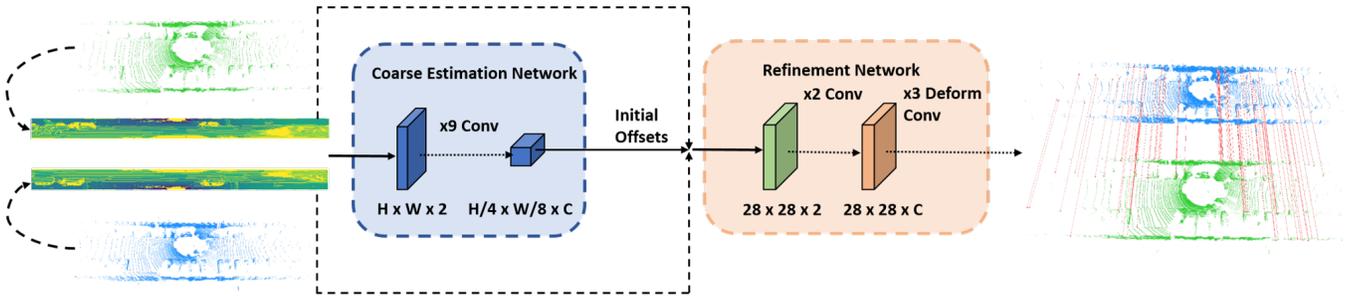

Fig. 2. Overview of our DMLO structure. Given two point clouds, we first encode them into cylinder images as inputs for network. And then DMLO predicts the correspondences in a coarse-to-fine manner. Then we convert the matching in cylinder image back to original 3D space, and then solve the relative motion between two scans.

## A. Overview

We depict our pipeline in Fig. 2. The full odometry system consists of frame by frame relative pose estimation, and various post-processing steps.

Different from previous work [14], [41], [42] which directly outputs the relative motion between two consecutive frames by CNN, we propose a geometric sound method for relative motion estimation. In particular, we first encode the two consecutive scans ($S_{t-1}$, $S_t$) using cylinder encoding method (Sec. III-B), then extract corresponding matching using CNN (Sec. III-C). Next, the extracted matching points are screened by Non-Maximum Suppression(NMS) and RANSAC method to reduce redundancy and outliers(Sec. III-D). Lastly, given the matching inliers, the relative pose between frames can be easily solved via standard SVD technique(Sec. III-D).

## B. Cylinder Encoding

Point cloud from LiDAR is typically represented by a set of sparse and unordered 3D points. Its unstructured nature makes processing LiDAR data more challenging than 2D images. Nevertheless, different from general 3D point cloud, the LiDAR data in one frame is actually one snapshot from one single view (a.k.a 2.5D data). This property leads us to design a more compact and representation for LiDAR point cloud than popular 3D convolution [43] or PointNet based methods [44], [45]. We follow [46], [14] to project it to a cylinder plane which renders the points to an ordered 2D grids with regular spacing. This structured form can be easily dealt with traditional CNN. Specifically, given the horizontal and vertical angular resolution ($\Delta\theta, \Delta\phi$), a 3D point $\boldsymbol{p} = (x, y, z)$ in Cartesian coordinates can be projected to grid $\boldsymbol{g}_p = (h, w)$ in the cylinder image, where

$$\begin{aligned} w &= \arctan2(y, x)/\Delta\theta, \\ h &= \arctan2(z, \sqrt{x^2+y^2})/\Delta\phi. \end{aligned} \quad (1)$$

Due to the noise of LiDAR, there may be more than one point in one grid. If this happens, we will keep the closest point. For each grid $\boldsymbol{g}$, we build two dimension features for it: the point range $r = \sqrt{x^2+y^2+z^2}$ and laser reflectivity value returned. By this efficient encoding method, we convert the sparse point cloud matching problem in 3D space into a (almost) dense matching problem in 2D image.

## C. Learning to Match Point Cloud by CNN

*a) Feature Embedding Network:* As shown in Fig. 2, our feature embedding network is a Siamese fully convolutional network with input size $H \times W \times 2$ and output size $\frac{H}{n} \times \frac{W}{m} \times C$, where $n, m$ are the network strides in vertical and horizontal directions, respectively and $C$ is the length of output features. To accommodate large motion prediction, a large maximum displacement $D$ is needed. Thus we design a coarse-to-fine cascade network structure as in optical flow estimation. For the first coarse estimation network, we use 9 layers of plain $3 \times 3$ convolution, and set $n = 4$, $m = 8$ and $D = 10$. It corresponds to a region of $40 \times 80$ in the original cylinder image, which is enough for large motion presented in odometry estimation. In the second refinement network, we use 2 layers of plain $3 \times 3$ convolution and 3 layers of $3 \times 3$ deformable convolution [47] to better fit the adaptive receptive field needs for matching. Also we set $n = m = 1$ and $D = 5$ in the refinement network to preserve the finest information.

As usual, we insert Batch Normalization (BN) [48] and Leaky ReLU [49] layers between each convolution layer. For each position in the resulted feature map, it represents the feature of $n \times m$ patch in the original cylinder encoded image. To calculate the cosine similarity, we first normalize the feature $\boldsymbol{x}_{i,j}$ as $\hat{\boldsymbol{x}}_{i,j} = \boldsymbol{x}_{i,j}/\|\boldsymbol{x}_{i,j}\|_2$. After that, a cross-correlation layer [50] of range $D \times D$ is applied. We modify the correlation layer can accept an additional offset $(u, v)$ as input so that it can be used the origin for the calculation of the similarity matrix. Specifically, for $\boldsymbol{x}_{i,j}^r$ is feature of $i$-th row and $j$-th column in reference frame, its similarity matrix $\boldsymbol{M}_{i,j}^{u,v} \in \mathbb{R}^{D \times D}$ is calculated through

$$\boldsymbol{M}_{i,j}^{u,v}(h, w) = (\boldsymbol{x}_{i,j}^r)^T \cdot \boldsymbol{x}_{i+u+h,j+v+w}^t \quad (2)$$

where $h \in [-\frac{D-1}{2}, \frac{D-1}{2}]$, $w \in [-\frac{D-1}{2}, \frac{D-1}{2}]$ and $\boldsymbol{x}_{i,j}^t$ is feature at grid $(i, j)$ in the target frame.

Finally, we obtain a similarity matrix of size $\frac{H}{n} \times \frac{W}{m} \times D \times D$. Based on the similarity matrix, we calculate the matching offset for each position as elaborated in the following section. The initial offsets output in the first coarse stage is used as the origin of correlation for the second refinement stage to enlarge the maximum predictable displacement while keeping the computational cost affordable.

*b) Matching Probability:* In contrast to traditional keypoint descriptors which focus on global matching across large pose variation. Odometry task only needs matching within a relatively small local region. Consequently, we apply a softmax transformation on the similarity matrix $\mathbf{M}_{i,j}^{u,v}$:

$$\mathbf{Q}_{i,j}^{u,v}(h,w) = \text{softmax}\left(\mathbf{M}_{i,j}^{u,v}(h,w)\right). \quad (3)$$

By introducing the competition among neighborhood areas, our matching method could fully utilize the hard examples near the ground-truth, which is beneficial for improving the accuracy of matching. Moreover, this makes our method can not only handle featureless area, but also could assign low confidence to the area of distinct but repetitive and ambiguous patterns, which is crucial to accurate confidence estimation.

*c) Ground Truth Generation:* For training, we only have frame level 6-DoF tranformation matrix $\mathbf{T}$. We need to transform it into pixel level correspondence for training. Firstly, for a grid $(h, w)$ with range value $\mathbf{D}(h, w)$, we define the inverse operation of Eqn. 1 as

$$\begin{aligned} z &= \mathbf{D}(h,w) \cdot \sin(h\Delta\phi) \\ x &= \mathbf{D}(h,w) \cdot \cos(h\Delta\phi) \cdot \cos(w\Delta\theta) \\ y &= \mathbf{D}(h,w) \cdot \cos(h\Delta\phi) \cdot \sin(w\Delta\theta) \end{aligned} \quad (4)$$

For brevity, we use $\Phi(\cdot)$ to represent Eqn. 1 and $\Phi^{-1}(\cdot)$ for Eqn. 4, and assume all 3D points coordinates as homogeneous. Given a point $\boldsymbol{p}_r$ in reference frame, we can regenerate $\boldsymbol{p}'_r$ by

$$\begin{aligned} \boldsymbol{c}_r &= \Phi(\boldsymbol{p}_r) & (5) \\ \boldsymbol{p}'_r &= \Phi^{-1}(\boldsymbol{c}_r) & (6) \\ \boldsymbol{p}_t &= \mathbf{T}\boldsymbol{p}'_r & (7) \\ \boldsymbol{c}_t &= \Phi(\boldsymbol{p}_t), & (8) \end{aligned}$$

where $\boldsymbol{p}_r, \boldsymbol{c}_r$ are 3D point and its projected grid index in reference scan, and $\boldsymbol{p}_t, \boldsymbol{c}_t$ are the correspondence in the target frame. Then correspondence $(\boldsymbol{c}_r, \boldsymbol{c}_t)$ is used as training pairs in the cylinder image. For every $\boldsymbol{c}_r$, our network predicts $\hat{\boldsymbol{c}}_t$. Then we use $(\boldsymbol{p}'_r, \hat{\boldsymbol{p}}_t)$ to solve the transformation matrix between two scans.

Cylinder encoding described in Eqn. 1 actually uses the center coordinates of a grid and range value in this grid to represent the corresponding points in 3D space. This will introduce inevitable quantization error since the point can not always fall at the center of the grid. To reduce this quantization error, we make an assumption that the range within one grid remains constant. We could create a virtual point $\boldsymbol{p}'_r$ at the center of the grid (Eqn. 5 and 6). Then we use the provided transformation matrix to transform it into target scan (Eqn. 7) and project to cylinder image (Eqn. 8). However, in some grid the range may greatly vary, thus this assumption does not hold. The training process may be affected by these noisy ground-truth. To remove these potential inaccurate ground-truth, we mask out the grids which have large range difference between the actual point in target frame and the projected virtual point $e = \mathbf{D}_t(\boldsymbol{c}_t) - \|\boldsymbol{p}_t\|_2$.

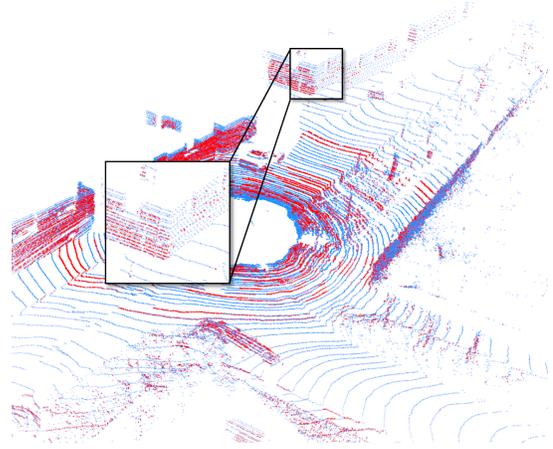

Fig. 3. We show the remaining points(red) which has less range difference in the same cylinder grid. The enlarged area shows that we eliminate the wall parallel to the beams while retaining the other side.

we only keep pairs $(\boldsymbol{c}_r, \boldsymbol{c}_t)$ with $e < 0.1(m)$ and remove the others. As shown in Fig. 3, we actually mask the points with small incident angles such as points on ground plane far away from LiDAR, since a small perturbation may result in large range error in these grids. By this way, we could guarantee that the ground-truth used for training is accurate with bounded quantization error.

*d) Sub-pixel Loss and Confidence Estimation:* As discussed in the previous paragraph, the projected grid $\boldsymbol{c}_t$ of the created virtual point may not be a integer. In this section, we describe our sub-pixel prediction technique to tackle this issue. For training, a bilinear sub-pixel loss [51] is adopted to transform a correspondence ground truth to distribution supervision. Let $(x, y)$ be grid index in target frame, $(\bar{x}_{i,j}, \bar{y}_{i,j})$ be ground-truth index for $(i, j)$ grid in reference frame and $\Delta x = |\bar{x}_{i,j} - x|, \Delta y = |\bar{y}_{i,j} - y|$, then the ground-truth probability of grid $(x, y)$, $\mathbf{P}_{i,j}(x, y)$ is:

$$\mathbf{P}_{i,j}(x,y) = \begin{cases} (1-\Delta x)(1-\Delta y) & \Delta x < 1 \text{ or } \Delta y < 1 \\ 0, & \Delta x \geqslant 1 \text{ and } \Delta y \geqslant 1 \end{cases} \quad (9)$$

Then the final sub-pixel loss is the cross entropy between the predicted probability $\mathbf{Q}_{i,j}$ and this ground-truth distribution $\mathbf{P}_{i,j}$:

$$\mathcal{L} = -\sum_{i,j}\sum_{x,y} \mathbf{P}_{i,j}(x,y) \log(\mathbf{Q}_{i,j}(x,y)) \quad (10)$$

As shown in Eqn. 9, for any $(\bar{x}_{i,j}, \bar{y}_{i,j})$, only one $2 \times 2$ window should has non-zero probability. During inference, we sum over each $2 \times 2$ window within the displacement range, and treat the window $\boldsymbol{W}^*_{i,j}$ with maximum $\mathbf{Q}^*_{i,j} = \sum_{(x,y)\in W_{i,j}} \mathbf{Q}_{i,j}(x,y)$ as the predicted window. We first normalize it by $\hat{\mathbf{Q}}_{i,j}(x,y) = \mathbf{Q}_{i,j}(x,y)/\mathbf{Q}^*_{i,j}$. Then the final predicted offsets $\hat{x}, \hat{y}$ are defined as:

$$\begin{aligned} \hat{x} &= \sum_{(x,y)\in \boldsymbol{W}_{i,j}} x \cdot \hat{\mathbf{Q}}_{i,j}(x,y), \\ \hat{y} &= \sum_{(x,y)\in \boldsymbol{W}_{i,j}} y \cdot \hat{\mathbf{Q}}_{i,j}(x,y). \end{aligned} \quad (11)$$

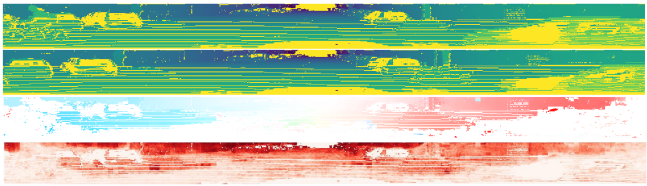

Fig. 4. Visualization of our correspondences and confidence. Top to bottom: cylinder image (reference frame), cylinder image (target frame), flow map and confidence map.

We further treat $\mathbf{Q}^*_{i,j}$ as the matching confidence for the grid $(i, j)$ in reference frame.

*D. From Initial Matching to Relative Motion Estimation*

*a) Non-Maximal Suppression (NMS):* We only need high precision and spatial uniformly distributed correspondences to solve a 6-DoF pose in 3D space. The original matchings from CNN are redundant and spatially clustered. Fortunately, as shown in Fig. 4, we have confidence estimation for each matching pair. These estimated confidence guides us to select good pairs for relative motion estimation. Therefore, we apply non-maximal suppression(NMS) over a fixed radius $r$ around each point. Namely, we sort the matching pairs by their confidence, and then iteratively find the matching with the highest confidence and surppress the other matchings within radius $r$. At last, we keep the top $N$ pairs $(\boldsymbol{p}'_r, \hat{\boldsymbol{p}}_t)$ with the highest matching scores.

*b) Rigid Alignment Problem in 3D Space:* After getting the matching pairs, the problem becomes how to solve the rigid transformation between two scans. Let $\mathcal{P} = \{\boldsymbol{p}_1, \boldsymbol{p}_2, ..., \boldsymbol{p}_N\} \subset \mathbb{R}^3$ and $\mathcal{Q} = \{\boldsymbol{q}_1, \boldsymbol{q}_2, ..., \boldsymbol{q}_N\} \subset \mathbb{R}^3$ denote two matched point sets. We denote the 6-DoF transformation matrix as $\mathbf{T} = \{\mathbf{R}, \boldsymbol{t} \mid \mathbf{R} \in \mathrm{SO}(3), \boldsymbol{t} \in \mathbb{R}^3\}$. The objective function can be formulated as:

$$\mathbf{T}^* = \underset{\mathbf{T}}{\arg\min} \frac{1}{N} \sum_{i=1}^{N} \|\boldsymbol{q}_i - \mathbf{T}\boldsymbol{p}_i\|_2^2. \tag{12}$$

Fortunately, the solution to this problem $\mathbf{T}^*$ can be given in closed form by Singular Value Decomposition (SVD) [52] as

$$\mathbf{R}^* = \mathbf{V}\mathbf{U}^T, \quad \boldsymbol{t}^* = -\mathbf{R}^*\overline{\boldsymbol{p}} + \overline{\boldsymbol{q}}, \tag{13}$$

where

$$\overline{\boldsymbol{p}} = \frac{1}{N}\sum_{i=1}^{N}\boldsymbol{p}_i, \quad \overline{\boldsymbol{q}} = \frac{1}{N}\sum_{i=1}^{N}\boldsymbol{q}_i,$$
$$\mathbf{H} = \sum_{i=1}^{N}(\boldsymbol{p}_i - \overline{\boldsymbol{p}})(\boldsymbol{q}_i - \overline{\boldsymbol{q}}), \tag{14}$$

and the SVD of $\mathbf{H}$ is denoted as $\mathbf{H} = \mathbf{U}\mathbf{S}\mathbf{V}^T$. As a common practice, RANSAC is used to reject outliers or the matchings on non-stationary objects.

*c) Other Components:* As a complete odometry system, we use a constant velocity model to predict ego-motion when there are not enough inliers after RANSAC. Also RangeNet++ [53] is used to remove moving objects to reduce the adverse effect of moving objects in ego-motion estimation. Finally, we maintain a local submap accumulated by previous scans and refine the pose by a scan-to-map NDT optionally.

## IV. EXPERIMENTS

In this section, we report the results of our DMLO method. We will briefly introduce the datasets we use first, then followed by odometry results and registration results under large motion. Finally, we conduct ablation studies to demonstrate the impact of each component in our system.

*A. Datasets*

*a) KITTI:* The KITTI odometry benchmark [54], [15] provides 4 camera images, Velodyne HDL-64E LiDAR scans, and 6-DoF ground truth pose from a high accuracy integrated GPS/IMU system. The dataset was recorded at 10fps when driving in highways, country roads and urban areas. It is the most widely used LiDAR odometry benchmark in the literature.

*b) Argoverse:* Argoverse 3D tracking dataset [16] is a recently proposed dataset tailored for autonomous driving. It contains 360-degree images from 7 cameras, forward-facing stereo cameras, 3D point clouds from LiDAR and 6-DoF ego-localization for each timestamp. The whole dataset consists of 100 sequences varying in length from 15 to 60 seconds collected in two US cities with more moving objects than KITTI. Different from KITTI, the point cloud is captured with two roof-mounted VLP-32 LiDAR sensors with an overlapping $40°$ vertical field of view. We use this dataset to demonstrate the robustness of our method to different hardware configurations.

*B. Implementation Details*

In KITTI dataset, we use the resolution from Velodyne HDL-64 specs and set $\Delta\phi = 0.4°, \Delta\theta = 0.2°$ in Eqn. 1. So that the size of cylinder frames is $H = 68, W = 1801$ with $26.9°, 360°$ FOV in vertical, horizontal. For network inputs, we crop it to $64 \times 1792$ from the center because of the sparsity in the edge. In Argoverse dataset, since the two overlapping VLP-32 LiDAR make the distribution of point cloud denser in middle areas and sparser on both sides, we set a smaller vertical resolution as $\Delta\phi = 0.3°$ within $20°$ FOV while keeping the horizontal resolution as $\Delta\theta = 0.2°$.

During training, we first sample a set of three successive scans, and for each three scans, we use it to form three training pairs $[S_{t-1}, S_t], [S_t, S_{t+1}]$ and $[S_{t-1}, S_{t+1}]$ as training samples. The reason for forming the last one by skipping one scan is to augment the training with large displacement. All models are trained using SGD with momentum of 0.9 for 60 epochs. The learning rate is set to 0.0005 at beginning and multiplied by 0.1 every 20 epochs with $10^{-5}$ weight decay. The batch size is 32 and 8 NVIDIA 1080Ti GPU is used to train the models with synchronized batch normalization (SyncBN) [57]. The whole framework is implemented with the MXNet library [58]. For inference, we set $r = 1$ in NMS to keep the correspondences spatial uniformly distributed. And the inlier threshold in RANSAC is 0.1m. If the inliners correspondences are less than 10, we will use the result from motion model.

TABLE I

ODOMETRY RESULTS ON KITTI ODOMETRY DATASET [54]. HERE $t_{rel}$ IS TRANSLATION ERRORS(%) ON ALL POSSIBLE SUBSEQUENCES OF 100M-800M. $r_{rel}$ IS ROTATIONAL ERROR($°$/100M) ON THE LENGTH OF 100M-800M.

| Seq | ICP-po2po $t_{rel}$ | ICP-po2po $r_{rel}$ | ICP-po2pl $t_{rel}$ | ICP-po2pl $r_{rel}$ | GICP [4] $t_{rel}$ | GICP [4] $r_{rel}$ | CLS [55] $t_{rel}$ | CLS [55] $r_{rel}$ | SUMA [56] $t_{rel}$ | SUMA [56] $r_{rel}$ | LOAM [2] $t_{rel}$ | LOAM [2] $r_{rel}$ | Ours $t_{rel}$ | Ours $r_{rel}$ | Ours+Mapping $t_{rel}$ | Ours+Mapping $r_{rel}$ |
|---|---|---|---|---|---|---|---|---|---|---|---|---|---|---|---|---|
| 00 | 6.88 | 2.99 | 3.80 | 1.73 | 1.29 | 0.64 | 2.11 | 0.95 | **0.68** | **0.23** | 0.78 | 0.53 | 0.83 | 0.44 | 0.73 | 0.44 |
| 01 | 11.21 | 2.58 | 13.53 | 2.58 | 4.39 | 0.91 | 4.22 | 1.05 | 1.70 | **0.54** | **1.43** | 0.55 | 3.14 | 1.15 | 1.91 | **0.54** |
| 02 | 8.21 | 3.39 | 9.00 | 2.74 | 2.53 | 0.77 | 2.29 | 0.86 | 1.20 | **0.48** | **0.92** | 0.55 | 1.08 | 0.53 | 1.01 | 0.58 |
| 03 | 11.07 | 5.05 | 2.72 | 1.63 | 1.68 | 1.08 | 1.63 | 1.09 | **0.74** | 0.50 | 0.86 | 0.65 | 0.80 | **0.49** | 1.06 | 0.53 |
| 04 | 6.64 | 4.02 | 2.96 | 2.58 | 3.76 | 1.07 | 1.59 | 0.71 | **0.44** | **0.27** | 0.71 | 0.50 | 0.71 | 0.53 | 0.61 | 0.66 |
| 05 | 3.97 | 1.93 | 2.29 | 1.08 | 1.02 | 0.54 | 1.98 | 0.92 | **0.43** | **0.20** | 0.57 | 0.38 | 0.90 | 0.46 | 0.65 | 0.33 |
| 06 | 1.95 | 1.59 | 1.77 | 1.00 | 0.92 | 0.46 | 0.92 | 0.46 | **0.54** | **0.30** | 0.65 | 0.39 | 0.91 | 0.46 | 0.65 | 0.34 |
| 07 | 5.17 | 3.35 | 1.55 | 1.42 | 0.64 | 0.45 | 1.04 | 0.73 | 0.74 | 0.54 | 0.63 | **0.50** | 0.58 | 0.51 | **0.53** | 0.51 |
| 08 | 10.04 | 4.93 | 4.42 | 2.14 | 1.58 | 0.75 | 2.14 | 1.05 | 1.20 | **0.38** | 1.12 | 0.44 | 1.02 | 0.48 | **0.93** | 0.48 |
| 09 | 6.93 | 2.89 | 3.95 | 1.71 | 1.97 | 0.77 | 1.95 | 0.92 | 0.62 | **0.22** | 0.77 | 0.48 | 1.02 | 0.45 | **0.58** | 0.30 |
| 10 | 8.91 | 4.74 | 6.13 | 2.60 | 1.31 | 0.62 | 3.46 | 1.28 | **0.72** | **0.32** | 0.79 | 0.57 | 1.08 | 0.59 | 0.75 | 0.52 |
| mean | 7.37 | 3.40 | 4.72 | 1.93 | 1.92 | 0.73 | 2.12 | 0.91 | **0.83** | **0.36** | 0.84 | 0.50 | 1.09 | 0.55 | 0.85 | 0.47 |

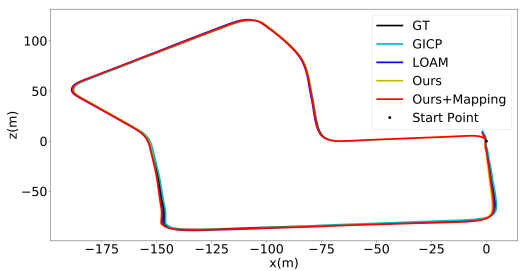

(a) sequence 07

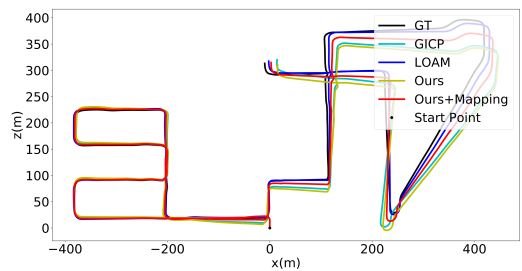

(b) sequence 08

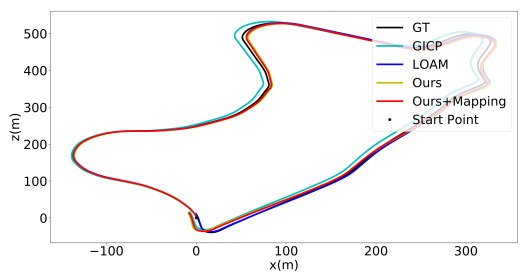

(c) sequence 09

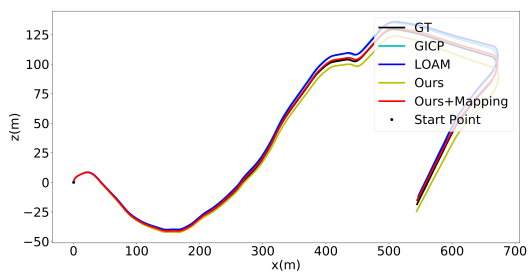

(d) sequence 10

Fig. 5. Qualitative results on the testing sequences of KITTI odometry dataset [54].

## C. Odometry Results

On KITTI dataset, we compare our methods with two kinds of odometry estimation methods. Geometry-based methods: point-to-point ICP (ICP-po2po), point-to-plane ICP (ICP-po2pl), GICP [4], CLS [55] and LOAM [2]; Learning-based methods: DeepLO [59], LO-Net [14] and Velas et al. [40]. All the methods are evaluated by official evaluation metrics. The three learning-based methods have different training/testing split, DeepLO uses 00-08/09-10, Velas uses 00-07/08-10, LO-Net uses 00-06/07-10 as training/testing, respectively. For fair comparison, we follow the setting of LO-Net which requires the least training data.

Tab. II shows testing results of all these learning-based networks, ours consistently outperforms the others by a large margin even with less training data. To fairly evaluate the performance of our DMLO in all sequences with other geometric based methods, we train our DMLO in leave-on-out manner. For example, to test sequence 00, we train our network on 01-10. We report both results of our method with and without backend mapping refinement. We compare the results with these methods in Tab. I. Compared with pure odometry method, our DMLO significantly outperforms the baselines ICP and GICP, while compared with other full SLAM systems, DMLO demonstrates competitive or better results. *Note that this is the first feature-based global method could compete with LOAM.* A qualitative results has been visualized in Fig. 5.

On the Argoverse dataset, we only compare with ICP-po2po, GICP and NDT, because LOAM is infeasible to apply

TABLE II

COMPARISON OF LEARNING-BASED METHODS IN KITTI [54]. WE ONLY SHOW THE RESULTS IN TESTING SPLIT.

| Seq. | [40] $t_{rel}$ | [40] $r_{rel}$ | DeepLO[59] $t_{rel}$ | DeepLO[59] $r_{rel}$ | LO-Net [14] $t_{rel}$ | LO-Net [14] $r_{rel}$ | Ours $t_{rel}$ | Ours $r_{rel}$ |
|---|---|---|---|---|---|---|---|---|
| 07 | - | - | - | - | 1.70 | 0.89 | **0.73** | **0.48** |
| 08 | 2.89 | - | - | - | 2.12 | 0.77 | **1.08** | **0.42** |
| 09 | 4.94 | - | 13.35 | 4.45 | 1.37 | 0.58 | **1.10** | **0.61** |
| 10 | 3.27 | - | 5.83 | 3.53 | 1.75 | 0.79 | **1.12** | **0.64** |

TABLE III

ODOMETRY EXPERIMENTS ON ARGOVERSE DATASET [16]. OUR DMLO OUTPERFORMS OTHER METHODS EVEN IN DIFFERENT HARDWARE(TWO VLP-32 LIDAR).

| Methods | RPE Mean | RPE Max | ATE Mean | ATE Max |
|---|---|---|---|---|
| ICP-Po2Po | 0.12 | 0.35 | 10.22 | 27.59 |
| G-ICP | 0.05 | 0.21 | 3.74 | 15.94 |
| NDT-P2D | 0.11 | 0.37 | 4.33 | 17.03 |
| Ours | **0.01** | **0.10** | **0.57** | **1.91** |

on two VLP-32 LiDAR. We evaluate the performance of all the methods on the official training/testing split. Since most sequences in Argoverse are shorter than 200 frames, the metrics from KITTI are not suitable. As [60] suggests, we use Absolute Trajectory Error(ATE) and Relative Pose Error(RPE) to evaluate the accuracy. Tab. III shows that our proposed method outperforms the others remarkably both in terms of ATE and RPE.

TABLE IV

REGISTRATION EXPERIMENTS IN KITTI [54].

| Methods | Angular Error(°) Mean | Max | Translation Error(m) Mean | Max |
|---|---|---|---|---|
| ICP-Po2Po | 0.139 | 1.176 | 0.089 | 2.017 |
| ICP-Po2Pl | 0.084 | 1.693 | 0.065 | 2.050 |
| G-ICP | 0.067 | 0.375 | 0.065 | 2.045 |
| AA-ICP | 0.145 | 1.406 | 0.088 | 2.020 |
| NDT-P2D | 0.101 | 4.369 | 0.071 | 2.000 |
| CPD | 0.461 | 5.076 | 0.804 | 7.301 |
| 3DFeat-Net | 0.199 | 2.428 | 0.116 | 4.972 |
| DeepVCP | 0.164 | 1.212 | 0.071 | **0.482** |
| Ours | **0.021** | **0.070** | **0.060** | 0.830 |

### D. Registration under Large Motion

Although our network is not designed to large scale registration, we also report the results comparing with recent proposed learning-based methods such 3DFeat-Net [35] and DeepVCP [34]. For registration, we simply change the max-displacement $D$ in coarse level from $10 \times 10$ to $40 \times 40$ in inference time without fine-tuning. Following DeepVCP, we sample the scans at 30 frame intervals and use all frames within 5m distance to it as registration target. The angular error and translation error is used to compare the performance of all methods. Given ground-truth transformation

TABLE V

END POINT ERROR(EPE) AND PERCENTAGE OF OUTLIERS AVERAGED OVER ALL GROUND TRUTH PIXELS(FL-ALL) IN KITTI 07-10 SEQUENCES. THE TABLE SHOWS THE DIFFERENT RESULTS OF FEATURE NORMALIZED(FN), SUBPIXEL(S) LOSS AND DEFORMABLE(D) CONV.

| Methods | Model | EPE(pixel) | Fl-all(%) |
|---|---|---|---|
| Basic | 5conv | 0.57 | 6.7 |
| Basic+FN | 5conv | 0.55 | 6.1 |
| Basic+FN+S | 5conv | 0.52 | 5.6 |
| Basic+FN+S | 8conv | 0.49 | 5.2 |
| Basic+FN+S | 2conv+3D | **0.45** | **4.2** |

TABLE VI

THE EFFECT OF MOTION MODEL (MM) AND MOVING OBJECTS REMOVAL (OR) ON KITTI. THE MEAN OF TRANSLATION AND ROTATION ERROR ARE COMPUTED AS TAB. I. MEAN† IS MEAN OF THE RESULTS IN THE TRAINING SET AND MEAN* IS THAT IN TESTING SET.

| Seq. | Basic $t_{rel}$ | $r_{rel}$ | Basic+MM $t_{rel}$ | $r_{rel}$ | Basic+MM+OR $t_{rel}$ | $r_{rel}$ |
|---|---|---|---|---|---|---|
| mean† | 1.66 | 0.59 | 1.24 | 0.58 | **1.16** | **0.57** |
| mean* | 0.97 | 0.54 | **0.96** | 0.57 | 1.00 | **0.53** |

$(\overline{\mathbf{R}}, \overline{\mathbf{t}})$ and prediction $(\mathbf{R}, \mathbf{t})$, the translation error is defined as $\|\mathbf{t} - \overline{\mathbf{t}}\|_2$ and angular error as $\theta = 2\arcsin(\frac{\|\mathbf{R} - \overline{\mathbf{R}}\|_F}{2\sqrt{2}})$. The results in Tab. IV indicate that, though we only use neighbor frames for training, our network can still generalize to distant registration task without any further tuning.

### E. Ablation Study

In this section, we conduct a set of ablation studies to understand how each component affects our performance. We first validate the efficacy of different network modules in the refinement network described in Sec. III-C. Then we experiment with and without motion model and moving objects removal. We utilize KITTI dataset in this section.

Tab. V shows that the Feature Normalized(FN), Subpixel(S) loss and Deformable(D) convolution significantly improve the accuracy of network. Specifically, the experiments between deeper network and deformable convolution clearly show that adaptive receptive field is essential for matching. Adopting deformable convolution is more economical than simply adding more plain convolution layers.

Tab. VI describes how the Motion Model(MM) and Moving Objects Removal(OR) affect our methods. As we can see, the influence of MM and OR on training set is greater than that on testing set. It is because sequence 01 in training set is recorded in highway with continuous featureless area and high speed. Tab. I also verifies that 01 is the most challenging sequence in all 11 sequences.

## V. CONCLUSIONS AND FUTURE WORK

This paper presents DMLO, a sparse matching LiDAR odometry framework. By deliberately designing the input representation and matching network, our method provides high accuracy correspondences which renders the feature-based global method comparable with local iterative methods for the first time. Comprehensive experiments on LiDAR odometry and registration tasks both demonstrate the effectiveness of our framework.

Besides the encouraging performance of learning-based method in LiDAR data, our framework is easy to generalize to other kinds of data such as RGBD and RGB+LiDAR. For future work, we will try to integrate more modalities. Especially, RGB data provides more dense and precious texture details than the point cloud data. We will investigate how to effectively fuse them into one matching framework.